\documentclass[twoside,11pt]{article}

\usepackage{jmlr2e}

\usepackage{amsmath}
\usepackage{subcaption}
\usepackage{booktabs}
\usepackage{multirow}
\usepackage{makecell}
\usepackage{float}

\usepackage{lastpage}
\jmlrheading{1}{2022}{1-\pageref{LastPage}}{6/22}{xx/yy}{farias00a}{Felipe Costa Farias, Teresa Bernarda Ludermir and Carmelo Jos\'e Albanez Bastos-Filho}

\ShortHeadings{Multi-criteria Machine Learning Model Selection}{Farias, Ludermir and Bastos-Filho}
\firstpageno{1}

\begin{document}

\title{Have we been Naive to Select Machine Learning Models? Noisy Data are here to Stay!}

\author{\name Felipe C. Farias \email felipefariax@gmail.com\\
    \name Teresa B. Ludermir \email tbl@cin.ufpe.br\\
    \addr Centro de Informatica\\
    Universidade Federal de Pernambuco\\
    Recife, PE, Brazil
    \AND
    \name Carmelo J. A. Bastos-Filho \email carmelofilho@ieee.org \\
    \addr E-Comp \\
    University of Pernambuco \\
    Recife, PE, Brazil}

\editor{}

\maketitle

\begin{abstract}
The model selection procedure is usually a single-criterion decision making in which we select the model that maximizes a specific metric in a specific set, such as the Validation set performance. We claim this is very naive and can perform poor selections of over-fitted models due to the over-searching phenomenon, which over-estimates the performance on that specific set. Futhermore, real world data contains noise that should not be ignored by the model selection procedure and must be taken into account when performing model selection. Also, we have defined four theoretical optimality conditions that we can pursue to better select the models and analyze them by using a multi-criteria decision-making algorithm (TOPSIS) that considers proxies to the optimality conditions to select reasonable models.
\end{abstract}

\begin{keywords}
neural networks, model selection, parallelization, over-fitting, over-searching, noise
\end{keywords}

\section{Introduction}




Machine Learning methods have been successfully applied to several areas because the algorithms can learn complex patterns from data. During the model development phase, we usually try to find the best model to solve a specific task. It is common to assess several different kinds of models and different combinations of hyper-parameters. After that, we need to select a model, or a set of models in the case of Ensembles, to deploy it to production to predict things in the real world. 

This paper investigates practical ways to select individual models to put in production for real-world usage. We call this problem model selection. The book \citep{hastie2009elements} states that the model selection goal is to estimate the performance of different models to choose the best one. One common approach is to use all the trained models as an ensemble. However, we are often interested in choosing a single model instead of several models to be used as an ensemble during operation. 

We can define over-fitting as a phenomenon that prevents the model from generalizing well on unseen data. In this case, the model can perfectly describe the Train set but have inferior performance at the Test set. In \citep{dietterich1995overfitting}, Dietterich claims that over-fitting is a phenomenon that emerges as we work too hard to find the best fit to the training data because there is a risk of fitting the noise by memorizing peculiarities of that training data instead of learning the general predictive rule.

In \citep{caruana2000overfitting}, the authors claim that MLP models with excess capacity (number of hidden neurons) generalize well if training with early stopping. An over-fitting overview was done in \citep{overfitting_overview_ying_2019} where the author discusses potential causes and solutions. To reduce the effects of over-fitting, they suggest the application of 4 perspectives to mitigate the over-fitting problem: (i) use of early stopping, (ii) network reduction, (iii) data expansion, and (iv) regularization. In our proposal, we are explicitly using the early stopping and regularization in the form of weight decay.

From the results of the paper \citep{zhang2021understanding}, one can confirm how powerful artificial Neural Networks (ANN) are and how important it is to use procedures to mitigate the over-fitting potential. They have found that ANN with sufficient parameters can perfectly fit even random labels. However, when using the trained model in the Validation set, the performance was terrible (as expected). The gap/disagreement between the random noise set and the correctly labeled set inspired our current proposal. 

If we analyze from the perspective of \citep{ng1997preventing}, the fact that real data is noisy makes model selection even harder. Consider an example in which we are performing model selection using a single-criterion approach to maximize the Holdout performance. Our Holdout set contains 100 samples, of which 80 samples are correctly labeled without noise, but 20 samples are noisy/wrong. Theoretically, suppose we have models A and B predicting 80\% of the time correctly. In that case, it might be the case that A is better than B in the Test set because A correctly predicted the 80/80 healthy and 0/20 problematic samples.
In comparison, B correctly predicted 60/80 healthy and 20/20 problematic samples, simply (over-)fitting the noise. Since we usually do not have the information of which points are correctly labeled or not. We can analyze the expectations. Now let us imagine that model A correctly predicted all the samples according to the current labeling (80/80 and 20/20) and B correctly predicted only 80\% (80/80 and 0/20), i.e., model A is over-estimated on the Holdout set. The most common approach is to select model A since it maximizes the single criterion. However, if we look into the Test set, model B will probably have a better performance because it does not fit the same noise present in the Holdout set. Since the over-estimated model would be selected due to noise in the Holdout set, we expect that it will correctly predict only 80\% of the correctly labeled data and present a bad performance on the noisy data in the test set. If model B is chosen, we would expect that it would correctly label roughly 100\% of the correctly labeled data and still have approximately lousy performance in the noisy portion of Test data. It is important to create strategies to better handle this phenomenom.

Although single-criterion is probably the most common approach to select machine learning models, there are works that tries to perform a multi-criteria model selection such as in \cite{ali2017accurate}, that proposed a multi-criteria decision making methodology using accuracy, time and consistency of each model to select the best one. It uses the TOPSIS \cite{topsis} ranking to measure the distance to the ideal classifier. The TOPSIS were also used in \citep{vazquezl2020framework} to model selection, but only used different performance metrics for a specific set. A fuzzy approach to work alongside with TOPSIS was proposed in \cite{akinsola2019performance}.
None of the mentioned multi-criterion proposals look into the training data to perform the model selection. We argue that the performance equilibrium between the individual sets can be a proxy measurement to the model robustness.


As far as we know, this is the first work investigating an immense number of NN architectures trained for several tabular datasets. The extensive regime of models allowed observing interesting behaviors during model selection that usually does not happen in small regime.

The objective of this paper is to compare the commonly used single-criterion against the multi-criterion model selection procedure. This paper is organized as follow. In Section~\ref{sec:methodology} we present our contribution and define optimality conditions of theoretical optimal models. In Section~\ref{sec:experiments} we explain the experimental setup. In Section~\ref{sec:results_and_discussion} we introduce and discuss the results. The conclusions are given in Section~\ref{sec:conclusions}. Finally, in Section~\ref{sec:future_works} we present future directions and opportunity of investigations on how to improve our contributions.

\section{Methodology}
\label{sec:methodology}
This section explains our methodology to study model selection on a large scale. The primary objective is to generate a partition of the data to analyze the effect of the model selection procedure in the model put in production.

\subsection{Data Splitting}
\begin{figure}[htbp]
\centering
\includegraphics[scale=0.47]{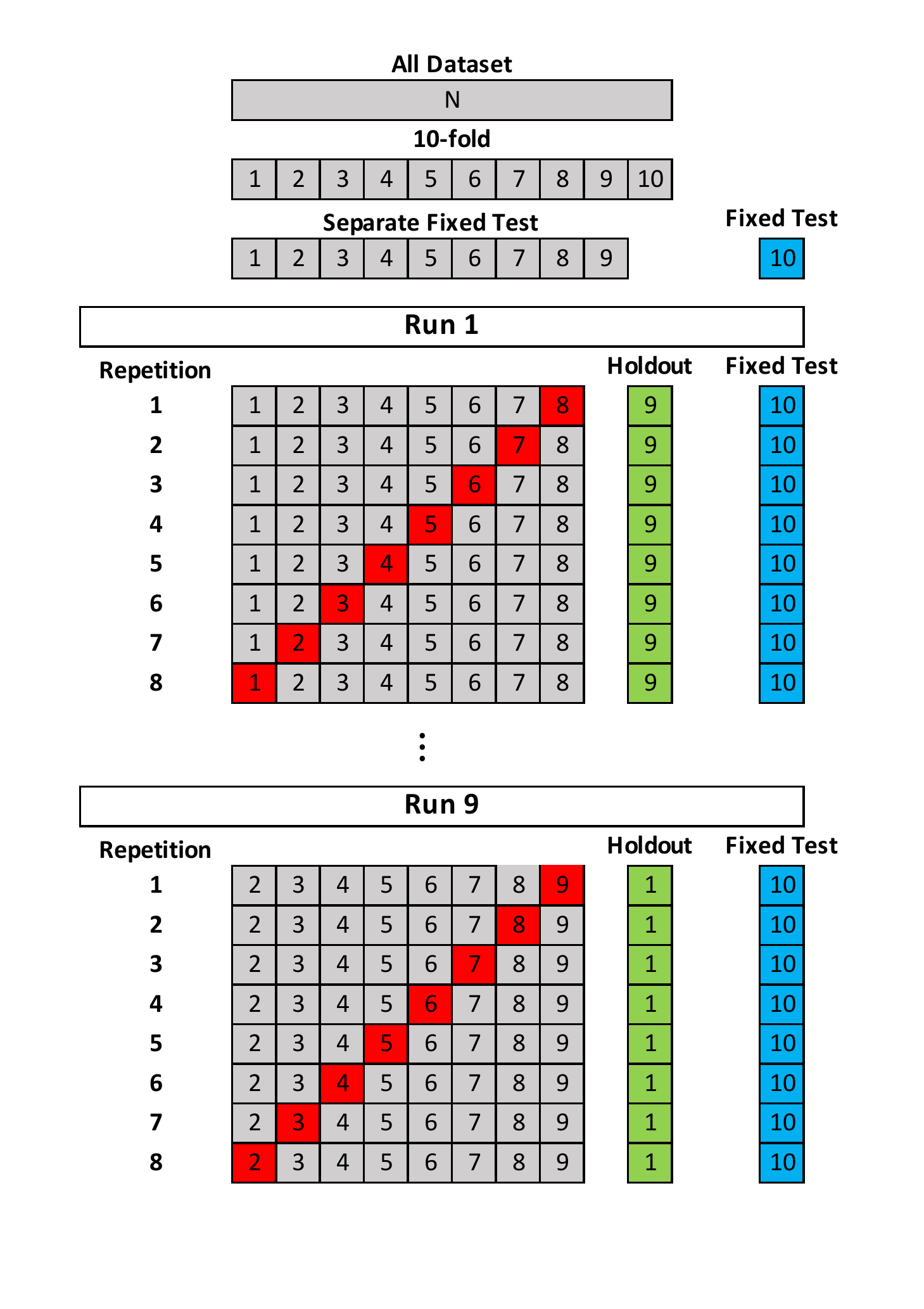}
\caption{Description of the sets constructions.}
\label{fig:splitting_procedure}
\end{figure}

The data splitting procedure is depicted in Figure~\ref{fig:splitting_procedure}. The first step is to divide the data into 10-folds. After that, the last fold is used as a Fixed Test set (blue box) among all the runs for a specific dataset. Another fold is used as a Holdout set (green box). Each of the remaining eight folds will be used once as a Validation set (red box), and the other seven will compose the Train set (gray box). It allows us to run nine times with different splits for the sets. We repeat the experiment twice, totaling 18 independent runs for each dataset.

It is worth mentioning that both Holdout and Fixed Test were never used during model training (neither was used in early stopping). Therefore, both sets can be seen as proxies for future unseen data, and can be used to estimate the models' performance when used in production.

\subsection{Models}
We have used the ParallelMLPs \citep{parallel_mlps} approach to create several different MLP architectures and analyze them simultaneously. The ParallelMLP was arranged to create architectures with 1 to 100 neurons in the hidden layer, using seven different activation functions (700 architectural possibilities) and eight repetitions (each repetition will use a different validation set). It creates a total of $100*7*8=5,600$ MLPs. We use the validation set to perform early stopping with the patience of 10 epochs.

With this strategy, seven splits are used as train splits. Since we have eight architectural repetitions (8 sub-networks are identical, they only differ by their random initialization), we round-robin the validation (early-stopping) split such that each one of the eight repetitions uses a different validation split. In order to accomplish this round-robin, we sample from training and validation splits (8 splits in total). We also use a train boolean mask that maps which dataset indices must be used as training samples for each sub-network to ignore those points during training dynamically. In practical terms, it means that the expected adequate batch size will proportionally have $7/8$ of the original batched samples, e.g., if $batch\_size=32$, it means that the expected number of training samples that each model will consider in each batch during back-propagation is $7/8*32=28$ samples, ignoring $1/8*32=4$ samples in expectation since it is marked for validation only. During the validation phase, we need to invert the training mask to obtain the validation mask. This procedure allows us to mimic the training of 8 independent trials varying which split is used to early stop the model. In our specific case, $700$ different architectures are being trained $8$ times, using a different validation set.

The model selection is a procedure responsible for selecting the best architecture, i.e., the one that maximizes the model performance on a specific dataset. The most common approach is using k-fold cross-validation and selecting the model with the best average performance on the validation set. In our understanding, this approach results in acceptable model choices, but it limits the learning process because the architecture is fixed. Consequentially, the number of parameters must be the same. 

As a Machine Learning system is composed by 4 components: (i) \textbf{model} -- including architectures -- to predict the targets; (ii) \textbf{loss function} to measure the success of the model; (iii) \textbf{optimization} algorithm to adjust the parameters of the model; and (iv) the \textbf{data}. If one of these components changes during the learning process, we cannot guarantee that we will have the same performance -- this is probably why k-Fold is a natural choice for ML hyper-parameter optimization. Even the initialization randomness can be a decisive factor for successfully learning a task once bad initial conditions difficult the learning dynamics of the model. Regarding the data, if we change the order in which the batches are chosen, we might have different learning dynamics. If we are performing a hyper-parameter search using a Tree or Gradient Boosting methods such as XGBOOST \citep{xgboost}, LightGBM \citep{lightgbm}, and even simple Random Forests \citep{random_forest}, we usually optimize the number of estimators. However, each estimator can grow to a specific depth, leading to different architectures and parameters even if we keep the same hyper-parameters. It is also observed when Support Vector Machines \citep{svm} (SVM) are used since the same hyper-parameters might create SVMs with different support vectors. This behavior of using a different number of parameters, even if we use the same hyper-parameters, mostly happens due to the learning dynamics given the initial conditions and the samples chosen to train the models, leading to different error landscapes, affecting the learning process. That is why we believe that forcing a NN to have a fixed number of parameters that will store the model's knowledge is not a good idea since the model's initialization and the splitting. Even the order of batches being presented will guide the model to different landscapes.

To the best of our knowledge, this is the first paper to investigate model selection using a large number of ANN. It was possible because we have used the ParallelMLPs algorithm to train several independent MLPs in an acceptable time window. Therefore, by looking into this regime, we could investigate further the relationship between different properties of NN trained on the same datasets. Due to the regime of a large number of trained models, we had the opportunity to investigate properties such as over-fitting and over-search on a large scale. \\

\subsection{Optimality Conditions}
We propose four optimality conditions based on desired properties that we believe the selected ANN models should have:
\begin{itemize}
    \item High Generalization Performance
    \item High Robustness
    \item Low Complexity
    \item No Premature Early Stopping \\
\end{itemize}

We argue that the more conditions a model meet, the more confident we are about its future performance when deployed in production to be used in the real world.

\subsubsection{High Generalization Performance}
How should one select a model from several possible candidates? Selecting the best Holdout model would be the best approach? 

When performing model selection, we are trying to collect the model that will perform the best during real-world operation. It means that we want a model with a high generalization. Therefore, the model should have good and close performance for the Holdout and Test sets.

Often only the Validation set is used to perform model selection. As we try to select the best model among several possibilities, we may incur into a problem if our objective is solely to maximize the Validation performance. In \citep{ng1997preventing} the authors suggest that selecting the ``apparent best" model is problematic because datasets are noisy, and the ``apparent best" model that maximizes the performance is learning the noise in the data. They suggested an approach to select models based on a percentile of the possibilities called \textit{percentile-cv}. However, it is a costly approach since it is based on a Leave-One-Out.

To analyze this optimality condition, we can define the Data Formation Rule Equation as 
\begin{equation}
\begin{split}
Y &= M(X) \\
\end{split}
\end{equation}
where the data ($Y$) can be explained by a prediction using the theoretical optimal Model ($M$) applied to the input data ($X$). However, real-world data very often contains noise, that's why we add the $\epsilon$ component to account for noise/aleatoric uncertainty \citep{hullermeier2021aleatoric}. 
\begin{equation} \label{eq:data_explanation}
\begin{split}
Y &= M(X) + \epsilon
\end{split}
\end{equation}
The noise can be related to the attribute or the targets/labels of the dataset. Therefore, for real-world data, we are usually influenced by $\epsilon > 0$ and  $\epsilon << M$, otherwise the dataset would probably contain too much noise to be helpful. For an interesting survey on dataset noise, one can read the \citep{survey_noise} work.
Since the $\epsilon$ is irreducible due to its stochastic behavior, and therefore the model should not explain, we should avoid the model to learn $\epsilon$. Consequently, if we have an incredible performance on a specific set of a dataset, it probably means that our model also has learned the noise of that set instead of learning what should be learned (model close to the theoretical optimal model). Usually, the more a model learns noise, the less generalizable it tends to be. We say that the single-criterion decision-making is naive because it assumes that the model should be able to explain the entire set, disregarding the actual irreducible error. When performing model selection, we really should be looking for a model that maximizes the metric performance of $Y$ due to the $M(X)$ and not the $\epsilon$ term. 

The NN must have the highest performance possible to solve a specific task when analyzing unseen data. The performance is usually measured only by looking into the Validation/Test set. Since we are using the Holdout and Test set as unseen data w.r.t. the Training and Validation sets, \textbf{the models must have the highest performance possible in those two sets}, on top of this assumption, we add that the metrics in both Holdout and Test sets should agree by having similar values, assuming they were pretty split without any biases. We have used a Stratified 10-fold and the Similarity-Based Stratified Splitting (SBSS) \citep{sbss}, both designed to achieve equivalent statistics for their folds. \\

\subsubsection{High Robustness}
\label{sec:optimality_condition_similar_sets_performance}
The Training and Validation set are one of the most important piece for the model to train, since the former updates the model's parameters and the latter is applied to early stopping in order to avoid over-fitting. Since they play such a relevant role, why do we ignore their information during the model selection procedure? One can argue that Training set performance will be highly overestimated because the parameters are being optimized directly to fit them. However, we have found that, inside the immense pool of MLP candidates that we were able to create, we still can find similar models w.r.t the Test set performance, but with the Train and Validation metrics close to each other that does not seem to be over-estimated as we would expect.

This condition is an extension of the previous optimality condition, but now looking into the training data (Train and Validation sets). It is helpful to analyze an extreme case in which we have the universe of points (all possible points) for a given task, such as all possible movements in a game. Suppose we train a model on this universe that perfectly fits the task (without over-fitting), finding the theoretical optimal model. In that case, the optimal decision boundaries were correctly found. Therefore we would have correctly predicted 100\% on all four sets (Train, Validation, Holdout, and Test). Suppose we randomly sub-sample the universe data and use only 80\% of the original data. In that case, the model could converge to the same parameters and still have perfect predictions on the four sets (which is very unlikely) or converge to a point where it can correctly predict 90\% of the cases. Since we randomly sub-sampled the data, we expect statistical properties to be preserved. Therefore, all four sets should have roughly 90\% of the time correct predictions. In other words, \textbf{all the Train, Validation, Holdout, and Test sets should have similar performances}. This equilibrium probably increases the robustness of the model and the trustfulness w.r.t. their estimated performance metrics.

Another way to analyze this condition is to consider two models, A and B, where the Train, Validation, Holdout, and Test metrics for model A is $0.99, 0.94, 0.92, 0.85$ and model B $0.86, 0.85, 0.85, 0.85$. Which one should be selected? Model B is probably more robust than A since model A presents some behaviors usually observed in over-fitted models (very high Training/Validation performance). Since we usually do not look into the Train and Validation metrics, we would probably select model A if only looking into the Validation/Holdout performance metric. That is why we defend the importance of considering more than a single-criterion to create better decisions.

When we perform model selection using a single set as a single-criterion decision given a large number of models, we are probably selecting a model that maximizes that metric by correctly predicting the corrected labeled points, but also ``correctly" predicting the noise on that data.

To better understand this optimality condition, we can decompose and specialize the Equation~\ref{eq:data_explanation} by each individual set as follows:

\begin{align}
Y_{train} &= M(X_{train}) + \epsilon_{train}\label{eq:data_explanation_decomposition_train} \\
Y_{validation} &= M(X_{train}) + \epsilon_{validation}\label{eq:data_explanation_decomposition_validation} \\
Y_{holdout} &= M(X_{train}) + \epsilon_{holdout}\label{eq:data_explanation_decomposition_holdout} \\
Y_{test} &= M(X_{train}) + \epsilon_{test}\label{eq:data_explanation_decomposition_test}
\end{align}

It is worth noticing that the model component $M$ is shared among all the sets since we are using the same model to explain the data.

We can define the model predictions as $\hat{Y} = M(X)=f(X|\theta)$ where $\theta$ are the parameters of the model. When training a NN on $Y_{train}$, we are trying to find the model's parameters $\theta$ such that $f(X_{train}|\theta)$ minimizes the loss function $L=loss(\hat{Y}_{train}, Y_{train} + \epsilon_{train})$. Specifically for the Train set we have:
\begin{equation}
    \theta = argmin \ loss(f(X_{train}|\theta), Y_{train} + \epsilon_{train})
    \label{eq:argmin_train}
\end{equation}


 Since the data explanation equations are explained by the model $M$ and the $\epsilon$ noise, there is a compromise between both terms. We theoretically should choose the model $M$ that explains (100\%) of set $Y$. However, in practice, each set $Y$ will have a percentage explained by $M$ and the remaining percentage by its own $\epsilon$ when we select the model $M$ with a single criterion that maximizes the performance metric or minimizes a loss function on a single set, e.g., the Train set -- explained by Equation~\ref{eq:data_explanation_decomposition_train}. It means that the selected model will probably incorporate the $\epsilon_{train}$ of that set, which is different from all the other $\epsilon$. This procedure will probably produce an over-fitted model. If we are willing to minimize the loss between $Y_{train}$ and $\hat{Y}_{train}$, we can state that $(Y_{train}-\hat{Y}_{train})\rightarrow0$, which is to say that $\hat{Y}=M(X_{train}) - \epsilon_{train}$. The model $M$ learned how to compensate for the noise $\epsilon_{train}$ of the training data. Therefore M can fully explain $Y_{train}$, also minimizing the loss between $Y_{train}$ and $\hat{Y}_{train}$ as can be seen in the following set of equations:
 
\begin{align}
    Y_{train} &= M(X_{train}) + \epsilon_{train}\\
    M(X_{train}) &= f(X_{train}|\theta) - \epsilon_{train}\\
    Y_{train} &= (f(X_{train}|\theta) - \epsilon_{train}) + \epsilon_{train} \\
    Y_{train} &= f(X_{train}|\theta)
\end{align}

The problem occurs when the previously learned model $M$ is applied to unseen data in the test set. The part less affected by $\epsilon_{test}$ in the Test set (explained mainly by the theoretical optimal model M term) will perform poorly due to the over-fitted model on the Training set, presenting a high generalization error as follows
\begin{align}
    Y_{test} &= M + \epsilon_{test} \\
    Y_{test} &= (f(X_{test} | \theta) - \epsilon_{train}) + \epsilon_{test}
\end{align}

To alleviate the problem of over-fitting, we usually use the Validation set constraining the model learning to minimize the loss function to explain the Train set in Equation~\ref{eq:subject_to_1} subjected to also minimize the loss in the Validation set simultaneously as in Equation~\ref{eq:subject_to_2}, but without using the Validation data to change the model's parameters.

\begin{align}
     \theta &= argmin \ loss(f(X_{train}|\theta), Y_{train} + \epsilon_{train})\label{eq:subject_to_1} \\
     \text{subject to:}\nonumber \\
     \theta &= argmin \ loss(f(X_{validation}|\theta), Y_{validation} + \epsilon_{validation})\label{eq:subject_to_2} 
\end{align}

When we perform model selection in a multi-criteria approach, e.g., maximizing all the sets, we are implicitly taking into account the noise on every set and trying to decrease the influence of each noise $\epsilon$ during the selection procedure such that we can find a model $M$ which mostly explains all the sets used in the multi-criteria decision. It will select models with worse results, but we claim that they are closer to reality since most of the explanation for each set comes from the shared model $M$ instead of a specific noise for a single set. In other words, we are trying to select the model that simultaneously optimizes the performance in the Training, Validation, and Holdout sets. We can understand it as finding M that minimizes $(Y_{train}-\hat{Y}_{train}), (Y_{validation}-\hat{Y}_{validation}), (Y_{holdout}-\hat{Y}_{holdout})$ simultaneously. In that case we are giving more importance to M term (since it is shared) relative to each set $\epsilon$: $M+\epsilon_{train}+M+\epsilon_{validation} + M\epsilon_{test} = 3M+\epsilon_{train}+\epsilon_{validation}+\epsilon_{test}$. That is why we believe the selected model when optimizing several sets is closer to the theoretical optimal model $M$ since it dilutes the influence of different $\epsilon$ across the sets.

The separation into several disjoint subsets to be presented to the NN during the learning process might also be an interesting research avenue to avoid over-fitting and improve model selection since we will probably decrease the influence of individual subset noises during the learning and selection process. However, we will leave that as future work.

\subsubsection{Low Complexity}
The complexity of the model is related to its representational capacity. When increasing the number of parameters in the model, it is natural to understand that it can store more information. However, we usually need more data to avoid over-fitting. If fewer parameters are used, less data is needed to train the model successfully. It is less prone to over-fit while decreasing the bias and increasing the difficulty for the model to fit the data. We can also analyze it through the lenses of the \textit{bias-variance tradeoff} \citep{hastie2009elements}. Generally, when model complexity increases, the variance increases, and the bias decreases. Ideally, the model should have low variance and low bias.
The best model should have the lowest complexity and the highest performance. It is usually hard to achieve due to the bias-variance trade-off. \textbf{The model should have the lowest number of neurons as possible}.\\

\subsubsection{No Premature Early Stopping}
Model initialization is very important to the training dynamics. We can analyze it using the same data, learning algorithm, and the number of epochs, only varying the random initialization of the model. We can divide the initialization into four categories: (i) the model starts in a sweet spot, and no further training is needed (early stopping on epoch 0) to predict the data (this is very unlikely) correctly; (ii) the model starts in a good region and a few epochs would be sufficient to optimize it, causing an early stopping in the beginning, because of the fast convergence due to a good initial guess (also unlikely); (iii) the model starts in a decent/moderate region and can be optimized for several epochs with late or no early stoppings; (iv) the model starts in an awful place that difficult the learning process causing early stoppings at the beginning of the training phase for various convergence issues such as vanishing/exploding gradients; The third case is more likely to happen since the training using Stochastic Gradient Descent algorithms tends to converge to a minimum due to the error decreasing (and therefore the gradients) throughout the number of epochs. 

If a model starts in a bad region (due to the random initialization), it might not be able to be adequately adjusted. Suppose we randomize a model and assess its performance in a dataset without the training phase. In that case, the chances that this model produces bad predictions is much larger than hitting a good spot and correctly predicting the data. Suppose we train using this bad randomly initialized model. In that case, it will probably stop training earlier than a good initialized model unless it randomly started very close to its convergence point in the parameters space (which is also unlikely). Therefore, we expect the \textbf{best model to be produced by a more extended training session (without premature early stoppings)}. \\

We believe that a multi-criteria model selection is better than a single-criterion because when we perform a multi-objective optimization, we usually will have a Pareto Front containing several non-dominated solutions. The solution that we will select using a multi-criteria method will probably not correspond to any extreme case in any dimension, which would be the natural choice in a single-criterion approach. Therefore, we expect that it alleviates the chance of selecting a model that is also fitting the noise of a specific set, consequently avoiding over-fitted models.

In order to select the best model among the 5,600 independently trained (in parallel) MLPs, we have analyzed several model selection policies related to the optimality conditions proposed in this paper.

\subsection{Selection Policies}
Here we describe the strategies we have used to select the models.

\subsubsection{Aggregation Policies}
To rank the models, we can aggregate them by the architecture, locally or globally, or treat them individually. In order to compare the aggregation at different levels, we proposed three aggregation policies to study each optimality condition.
Each aggregation group has different approaches on how to aggregate metrics in order to find the best architecture:

\begin{itemize}
    \item \textbf{Individual}: Rankings are treated as individual models' ranks in each run for each dataset, without any aggregation. For each run and dataset, we compare 5,600 individual models to find the one that maximizes the objectives the most.
    
    \item \textbf{Local}: Rankings are averaged by architecture (8 repetitions of the same architecture, defined by the number of neurons and activation function), containing a mutually exclusive validation set for early stoppings. For each run and dataset, we are comparing between $5,600/8=700$ different architectures.
    
    \item \textbf{Global}: In this aggregation, we averaged the rankings of each eight architectural repetitions of all the 18 runs, leading to $18*8=144$ architectures trained in different folds but sharing the same Fixed Test set. Once we find the best architecture given all the runs w.r.t. their rankings, we select the model among its eight architectural repetitions for each run. This strategy allows us to understand if we have a "best architecture" regarding a specific dataset and an applied ranking.
\end{itemize}

.
\subsubsection{Ranking Policies}
In order to rank models, we proposed several ranking policies that use different strategies to study our optimality conditions. 

Even though policies that use the Test metric should be avoided in real life because we would be selecting a model based on the Test set, it serves as an upper bound or an approximation of the theoretical optimal model to compare with and analyze the relationship between other policies that do not include the Test set metric as part of the decision criteria.

\begin{itemize}
    \item \textbf{Single Sets} \\ Here, we pick the models that maximize the performance metric on a specific set.
    \begin{itemize}
        \item \textbf{Train}: Selects the model with the best performance on the train set. Even though it is not a common approach, we can use it to study over-fitting.
    
        \item \textbf{Validation}: Selects the model with the best performance on their respective validation set that was used to early stop the model training.
        
        \item \textbf{Holdout}: Selects the model with the best performance on the holdout set.
        
        \item \textbf{Test}: Selects the model with the best performance on the Test set for comparison purposes.
    \end{itemize}
\end{itemize}

Local Validation is the most common approach, where the average performance of the same model architecture is calculated on the validation set. In our case, we calculate the average of the eight repetitions trained using a mutually exclusive validation set. Once we have defined the architecture, we select the model with the best performance on its validation set. 

We argue that ML models contain several important intrinsic aspects usually neglected when performing the model selection. Summarizing the model and its idiosyncrasies by naively treating model selection as a single-criterion decision task when we only look into a specific set performance metric such as Validation or Holdout accuracy is probably an oversimplified and sub-optimal approach.
Therefore, we propose that model selection should be a multi-criteria procedure task and use a combination of properties to rank the models. We have tried to maximize the \textit{performance metric} and the \textit{number of epochs}, and minimize the \textit{number of neurons}, simultaneously. 
We used Multi-Criteria Decision-Making (MCDM) algorithms \citep{mcdm, survey_mcdm,salabun2020mcda} to rank the models. Specifically, a straightforward and known algorithm is the Technique for Order of Preference by Similarity to Ideal Solution (TOPSIS) \citep{topsis}. In this algorithm, we will select the model that minimizes the geometric distance from the Positive Ideal Solution (PIS) and simultaneously maximize the geometric distance from the Negative Ideal Solution (NIS). We can understand the PIS and NIS as the best possible model (maximum metrics, one neuron, maximum stopped epoch) and the worst possible model (minimum metrics, 100 neurons, minimum stopped epoch), respectively. After ranking the models, we apply a Pareto Dominance criteria filter to remove dominated solutions and select the best-ranked model among the non-dominated Pareto solutions. We relied on the PyMCDM library \citep{pymcdm} to use the TOPSIS algorithm. 

Since the MCDM methods create a single scalar rank based on several variables, it will intrinsically incorporate a trade-off or compromise analysis during its execution. Consequentially, we will not be selecting the best performance for a single optimized variable, alleviating the issue of over-searching. The MCDM methods can use different weight scheme combinations and the vectors' normalization.

To improve the paper's readability, we are naming TOPSIS policies after the first letter of what we include in its multi-criteria.

\begin{itemize}
    \item \textbf{High Generalization Performance} \\ Both Holdout and Test performances should be maximized simultaneously.
    \begin{itemize}
        \item \textbf{THT}: TOPSIS maximizing Holdout and Test metrics.
    \end{itemize}

    \item \textbf{High Robustness} \\ The Train and Validation sets also need to be maximizes.    
    \begin{itemize}
        \item \textbf{TTVH}: TOPSIS maximizing Train, Validation, and Holdout metrics
        \item \textbf{TTVHT}: TOPSIS maximizing Train, Validation, Holdout, and Test metrics
    \end{itemize}

    \item \textbf{Low Complexity} \\ We additionally provide the number of neurons to be minimized in the TOPSIS.
    \begin{itemize}
        \item \textbf{TTVHN}: TOPSIS maximizing Train, Validation, Holdout, and minimizing the Number of Neurons.
        \item \textbf{TTVHTN}: TOPSIS maximizing Train, Validation, Holdout, Test, and minimizing the Number of Neurons.
    \end{itemize}

    \item \textbf{No Premature Early Stopping} \\ We additionally provide the Number of Epochs to be maximized (E) in the TOPSIS. For comparison purposes, we also tried to minimize the Number of Epochs (B -- Begin of the training).
    \begin{itemize}
        \item \textbf{TTVHNE}: TOPSIS maximizing Train, Validation, Holdout, Number of Epochs, and minimizing the Number of Neurons.
        \item \textbf{TTVHNB}: TOPSIS maximizing Train, Validation, Holdout, and minimizing the Number of Neurons and Number of Epochs (Begin of the train) to comparisons purposes.
        \item \textbf{TTVHTNE}: TOPSIS maximizing Train, Validation, Holdout, Test, Number of Epochs, and minimizing the Number of Neurons.
        \item \textbf{TTVHTNB}: TOPSIS maximizing Train, Validation, Holdout, Test, and minimizing the Number of Neurons and Number of Epochs (Begin of the train) for comparisons purposes.
    \end{itemize}
\end{itemize}



The policies are using the number of neurons as a tiebreaker (models with fewer neurons are preferred) in case models have the same metric being ordered.

As the splits are sampled so that the statistical characterization should be the same, we claim that the best model is not the model that only maximizes the Test set performance metric. However, it also needs a slight disagreement regarding the performance along the Train, Validation, and Holdout sets. If we select the model solely based on the Test performance, the model might be over-fitting in the Test set because its performance might be much higher than on the other sets. Therefore, we are considering good models that balance the performance on all the individual sets. 
We suggest using the TTVH policy if we want to maximize the accuracy and TTVHN policy to still have good accuracy but focusing on smaller models since both of the policies does not look into the Test set and the No Premature Early Stopping criteria seems to need further adjustments regarding the weight that this condition should be given.

\section{Experiments}
\label{sec:experiments}

\subsection{Computing Environment}
A Machine with 16GB RAM, 11GB NVIDIA GTX 1080 Ti, and an I7-8700K CPU @ 3.7GHz containing 12 threads were used to perform the simulations. All the code was written using PyTorch \citep{pytorch}.

\subsection{Number of Neurons}
We have created MLPs containing from 1 to 100 neurons in their hidden layer. It gives us 100 different architectures.

\subsection{Activation Functions}
We have used seven activation functions (Identity, GELU, LeakyReLU, ReLU, SeLU, Sigmoid, Tanh). Combined with the number of neuron variations, we get $100*7=700$ different architectures.

\subsection{Splitting Strategy}
We have assessed our proposal with the Similarity-Based Stratified Splitting (SBSS) \citep{sbss} and without it -- an ordinal stratified 10-fold -- with one fold being a fixed test split, another one a fixed holdout split, and the remaining eight splits being used as training and validation splits. We have used 10 splits because it is one of the most common k-fold setup used in machine learning.

\subsection{Datasets}

We assessed the proposed selection policies and optimality conditions in several situations, such as many features and labels, a low number of samples, and dataset imbalances. We have used 14 datasets from UCI~\citep{uci} listed in Table~\ref{tab:datasets}.
We calculated the Imbalance of each dataset by adapting the suggestion in \citep{imbalancemetric} according to Eq.~\ref{eq:dataset_imbalance}, resulting in 0 when the dataset is balanced and 1 otherwise. 

\begin{equation}
    \label{eq:dataset_imbalance}
    Imbalance = 1 - \frac{\sum_{i=1}^{k}\frac{c_i}{n}log(\frac{c_i}{n})}{log(k)}
\end{equation}
where $n$ is the number of samples; $k$ is the number of labels, and $c_i$ is the number of samples in label $i$.
\begin{table}[!htpb]
    \centering
    \resizebox{\columnwidth}{!}{
        \begin{tabular}{p{6cm}cccc}
            \toprule
            \textbf{Dataset}                 & \textbf{\# Features} & \textbf{\# Labels} & \textbf{\# Samples} & \textbf{Imbalance} \\
            \midrule
            balance-scale                    & 4                    & 3                  & 625                 & 0.17               \\
            blood-transfusion-service-center & 4                    & 2                  & 748                 & 0.21               \\
            car                              & 6                    & 4                  & 1728                & 0.40               \\
            diabetes                         & 8                    & 2                  & 768                 & 0.07               \\
            tic-tac-toe                      & 9                    & 2                  & 958                 & 0.07               \\
            ilpd                             & 10                   & 2                  & 583                 & 0.14               \\
            vowel                            & 12                   & 11                 & 990                 & 0.00               \\
            australian                       & 14                   & 2                  & 690                 & 0.01               \\
            climate-model-simulation-crashes & 18                   & 2                  & 540                 & 0.58               \\
            vehicle                          & 18                   & 4                  & 846                 & 0.00               \\
            credit-g                         & 20                   & 2                  & 1000                & 0.12               \\
            wdbc                             & 30                   & 2                  & 569                 & 0.05               \\
            ionosphere                       & 34                   & 2                  & 351                 & 0.06               \\
            satimage                         & 36                   & 6                  & 6430                & 0.04               \\
            libras move                     & 90                   & 15                 & 360                 & 0.00               \\
            lsvt                             & 310                  & 2                  & 126                 & 0.08               \\
            \bottomrule
        \end{tabular}
    }
    \caption{Datasets used as benchmarks}
    \label{tab:datasets}     
\end{table}

We selected datasets with different properties such as number of features, labels, samples and imbalance levels. This creates different challenges for each dataset that are useful to simulate several situations which our proposal may be exposed when working with real world data.

\section{Results and Discussion}
\label{sec:results_and_discussion}
In this section we present the results and discussions of our experiments.

\begin{figure}[!htpb]
\includegraphics[width=1\textwidth]{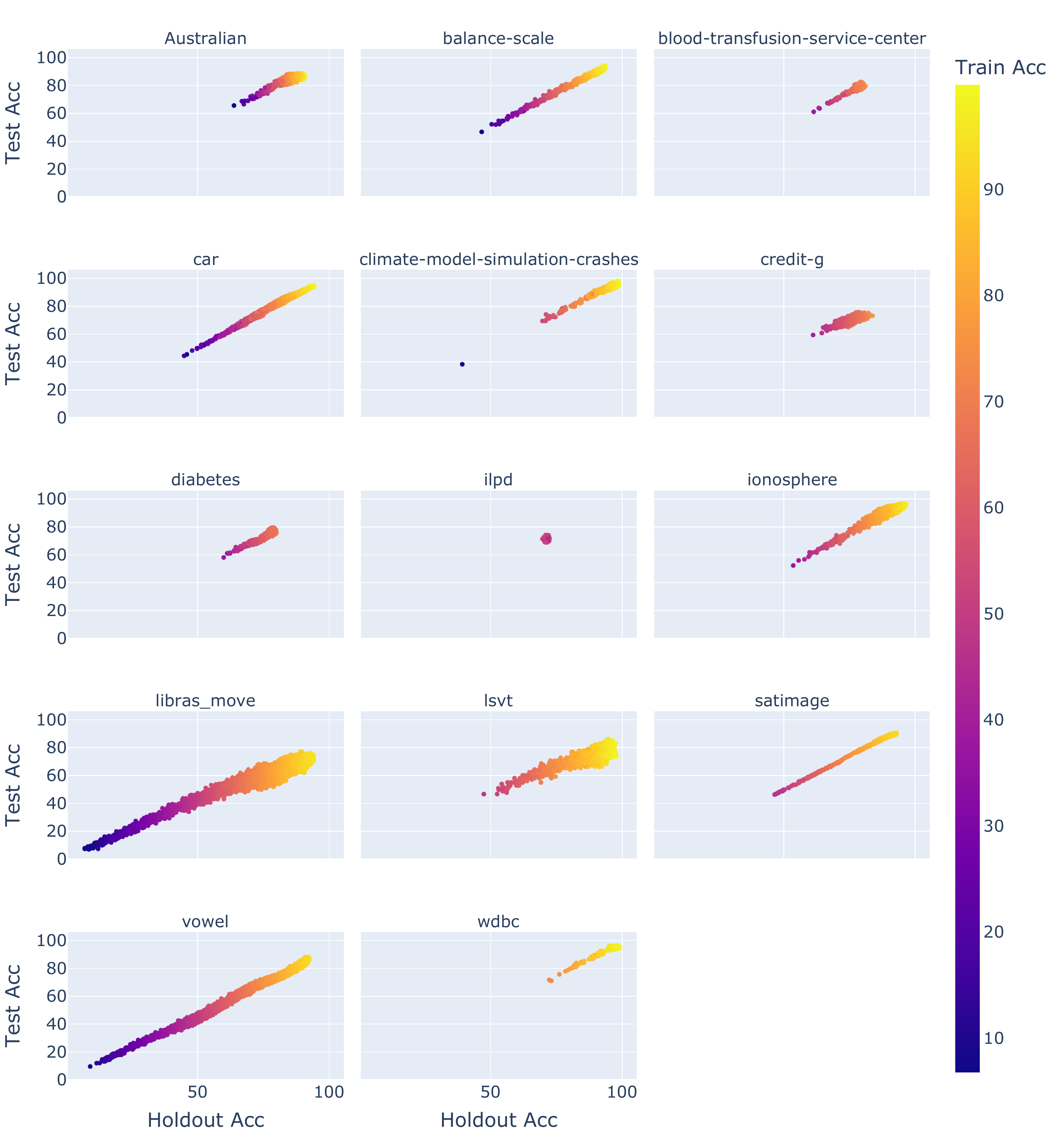}
\caption{Average Train, Holdout, and Test accuracies grouped by run, dataset, and architecture.}
\label{fig:average_group_run_dataset_architecture}
\end{figure}

Each point in Figure~\ref{fig:average_group_run_dataset_architecture} is the average of 8 models with the same architecture (combination of the number of neurons and activation function) in a specific run for each dataset. As expected, we can observe a high correlation in the distribution of Train, Holdout, and Test accuracies.

\begin{figure}[!htbp]
\centerline{\includegraphics[scale=0.8]{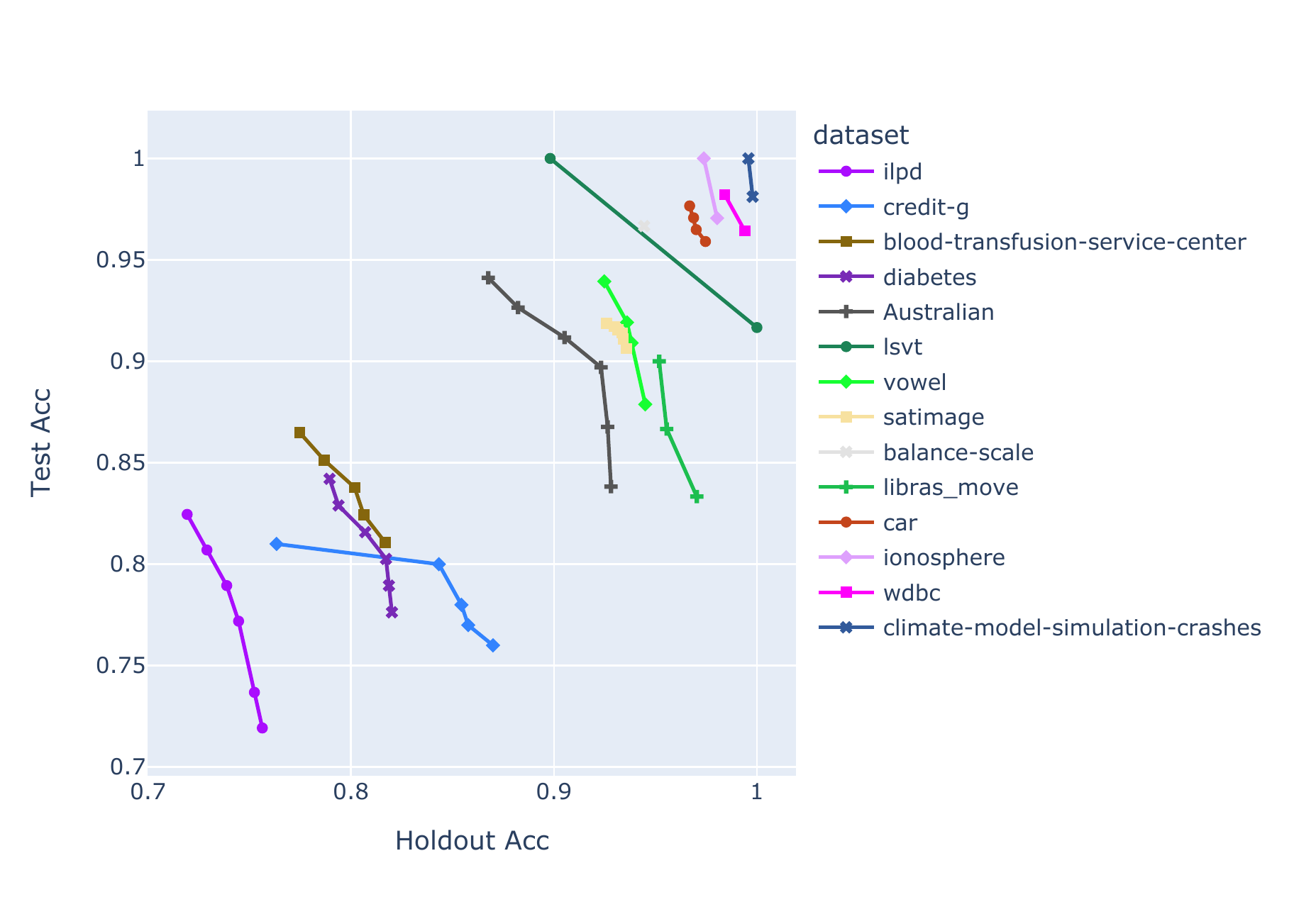}}
\caption{Pareto front for each dataset.}
\label{fig:pareto_fron_each_dataset}
\end{figure}

In Figure~\ref{fig:pareto_fron_each_dataset} we have created the Pareto Front using the Holdout and Test sets performances considering all the models for each dataset. Pareto Front is a concept that tries to capture trade-offs between variables being used in a multi-objective optimization process. Since we are using two variables in this plot (Holdout and Test accuracies), we create a set of models such that there are no other models that can improve some variable without reducing others. Neither the Holdout nor the Test set participates during the model training. Therefore, from the model standpoint, those two specific sets can be considered as two proxies of real data that the model will probably see during the real operation.

As we can see, if the model that performs best in the Holdout set is used, it is not guaranteed that the model will have the best performance in the Test set. Based on this finding, we can state that selecting a model that only maximizes the Holdout set will not give us the best model to be used as a production model.

\subsection{SBSS vs. Stratified 10-fold}

A comparison between a 10-fold SBSS and an ordinary Stratified 10-fold is presented in Table~\ref{tab:sbss_no_sbss}. It is easy to realize that the 10-fold SBSS consistently outperforms the ordinary Stratified 10-fold splitting. Therefore, the subsequent analysis will be done only using SBSS.

\begin{table}
\centering
\begin{tabular}{c|cccc}
\toprule
\multirow{2}{*}{Splitting Strategy}& \multicolumn{4}{c}{Accuracy $\uparrow$} \\
 &      Train & Validation &    Holdout &       Test \\
\midrule
10-fold SBSS               &  \makecell{83.62\\(12.72)} &  \makecell{83.68\\(11.24)} &  \makecell{84.28\\(11.29)} &  \makecell{80.97\\(12.45)} \\
Stratified 10-fold &  \makecell{73.94\\(16.62)} &  \makecell{76.63\\(14.72)} &  \makecell{76.48\\(14.42)} &  \makecell{74.87\\(15.73)} \\
\bottomrule
\end{tabular}
\caption{Accuracy average and standard deviation (in parenthesis) for all the $5,600$ generated models, $18$ runs and $14$ datasets, totaling $1,411,200$ models for each Splitting Strategy.}
\label{tab:sbss_no_sbss}

\end{table}

The following results tables contain the average and standard deviations in parenthesis for each Policy for the selected models w.r.t. the Number of Neurons (\# Neurons - lower is better), Number of trained Epochs due to Early Stopping (\# Epochs - higher is better), Accuracies for Train, Validation, Holdout, and Test sets (higher is better), and the Disagreement (smaller is better) as the average of the absolute individual difference between Train-Validation, Holdout-Test and All (averaged absolute difference for each combination of paired sets). Wilcoxon statistical significance test tables contain three symbols to represent: ($\blacktriangle$) row distribution significantly larger than column; ($\equiv$) row distribution not significantly different from the column; ($\triangledown$) row distribution significantly smaller than the column. If the variable is Accuracy, policies that produce larger values are better. On the other hand, if the variable is Disagreement, smaller values are better. The Summary contains the sum of $\blacktriangle (1)$, $\equiv (0)$, $\triangledown (-1)$ for each policy in rows.

\subsection{High Generalization Performance Results}
The results for each aggregation and ranking Policy are presented in Table~\ref{tab:high_generalization_performance} and are useful for understanding the High Generalization Performance condition. Also, the Wilcoxon Statistical Tests regarding the Test performance can be seen in Table~\ref{tab:statistical_high_generalization_performance}.

\begin{table}[htpb]
\centering
\resizebox{\columnwidth}{!}{
\begin{tabular}{cc|c|c|cccc|ccc}
\toprule
 \multicolumn{2}{c|}{\multirow{2}{*}{Policy}}& \multirow{2}{*}{\# Neurons $\downarrow$}   & \multirow{2}{*}{\# Epochs $\uparrow$} &\multicolumn{4}{c|}{Accuracy $\uparrow$} & \multicolumn{3}{c}{Disagreement $\downarrow$} \\
              &      &     &   &        Train &   Validation &      Holdout &         Test & Train-Validation & Holdout-Test &         All \\
\midrule
        \multirow{9}{*}{Individual} &      Train & \makecell{73.47\\(29.13)} &   \makecell{86.02\\(24.6)} &  \makecell{92.45\\(7.72)} &  \makecell{87.47\\(7.8)} & \makecell{91.28\\(7.71)} &  \makecell{85.06\\(9.5)} &      \makecell{5.03\\(4.21)} &   \makecell{6.4\\(6.84)} & \makecell{4.66\\(4.05)} \\
         & Validation & \makecell{47.02\\(34.84)} &  \makecell{62.99\\(35.21)} &   \makecell{87.85\\(8.8)} & \makecell{91.86\\(6.85)} & \makecell{88.97\\(7.98)} & \makecell{84.25\\(9.78)} &      \makecell{4.26\\(3.96)} &  \makecell{5.21\\(6.76)} & \makecell{4.38\\(3.83)} \\
         &    Holdout &  \makecell{76.41\\(27.5)} &  \makecell{84.79\\(25.13)} &  \makecell{92.08\\(7.92)} & \makecell{89.94\\(7.79)} & \makecell{91.76\\(7.66)} & \makecell{85.54\\(9.29)} &      \makecell{2.77\\(2.66)} &  \makecell{6.37\\(6.53)} & \makecell{3.94\\(3.51)} \\
         &       Test &  \makecell{42.8\\(33.91)} &  \makecell{63.72\\(34.72)} &   \makecell{87.42\\(8.8)} & \makecell{87.07\\(7.64)} & \makecell{87.68\\(7.98)} & \makecell{90.61\\(7.05)} &      \makecell{2.67\\(3.25)} &  \makecell{4.29\\(3.51)} & \makecell{3.13\\(2.41)} \\
         &        THT &  \makecell{75.5\\(24.61)} &  \makecell{78.21\\(27.19)} &  \makecell{91.22\\(8.36)} & \makecell{89.05\\(7.89)} &  \makecell{90.9\\(8.07)} & \makecell{89.65\\(7.34)} &      \makecell{2.83\\(2.67)} &  \makecell{3.52\\(3.26)} &  \makecell{2.67\\(2.0)} \\
         \midrule
             \multirow{9}{*}{Local} &      Train & \makecell{85.19\\(18.79)} &  \makecell{80.42\\(26.98)} &  \makecell{91.75\\(8.21)} &  \makecell{88.0\\(7.93)} &  \makecell{90.95\\(8.0)} & \makecell{85.21\\(9.43)} &       \makecell{3.86\\(3.6)} &  \makecell{5.98\\(7.27)} & \makecell{4.12\\(4.09)} \\
              & Validation & \makecell{80.45\\(20.88)} &  \makecell{62.89\\(33.46)} &   \makecell{89.07\\(8.9)} &  \makecell{90.8\\(7.08)} & \makecell{89.65\\(8.21)} & \makecell{85.14\\(9.38)} &      \makecell{2.68\\(2.69)} &   \makecell{4.94\\(6.1)} & \makecell{3.52\\(3.19)} \\
              &    Holdout &  \makecell{86.41\\(17.5)} &  \makecell{78.35\\(27.47)} &  \makecell{91.52\\(8.01)} &  \makecell{89.37\\(7.6)} &  \makecell{91.2\\(7.71)} &  \makecell{85.01\\(9.6)} &      \makecell{2.76\\(2.76)} &  \makecell{6.36\\(7.27)} & \makecell{3.94\\(3.97)} \\
              &       Test & \makecell{68.52\\(30.98)} &   \makecell{61.08\\(34.2)} &  \makecell{88.61\\(9.19)} & \makecell{88.13\\(7.81)} & \makecell{88.77\\(8.51)} & \makecell{89.05\\(7.49)} &      \makecell{2.28\\(2.62)} &  \makecell{3.46\\(3.28)} & \makecell{2.48\\(2.11)} \\
              &        THT & \makecell{82.75\\(19.23)} &  \makecell{68.67\\(30.14)} &  \makecell{90.56\\(8.51)} & \makecell{88.58\\(7.57)} & \makecell{90.28\\(8.03)} & \makecell{88.76\\(7.59)} &      \makecell{3.26\\(3.32)} &  \makecell{3.58\\(3.79)} &  \makecell{2.86\\(2.4)} \\
            \midrule
            \multirow{9}{*}{Global} &      Train & \makecell{91.86\\(11.95)} &  \makecell{75.19\\(27.65)} &  \makecell{91.43\\(8.26)} &  \makecell{87.85\\(7.9)} & \makecell{90.69\\(8.01)} & \makecell{85.34\\(9.22)} &       \makecell{3.73\\(3.9)} &  \makecell{5.55\\(6.55)} & \makecell{3.85\\(3.87)} \\
             & Validation & \makecell{92.29\\(13.52)} &  \makecell{61.85\\(29.62)} &  \makecell{89.17\\(8.28)} & \makecell{90.13\\(7.42)} & \makecell{89.51\\(7.95)} &  \makecell{84.91\\(9.8)} &      \makecell{1.96\\(1.89)} &  \makecell{5.19\\(6.77)} & \makecell{3.33\\(3.48)} \\
             &    Holdout & \makecell{87.36\\(23.97)} &  \makecell{75.85\\(26.96)} &  \makecell{90.99\\(8.93)} & \makecell{88.78\\(7.87)} & \makecell{90.83\\(8.02)} & \makecell{85.48\\(9.18)} &      \makecell{3.08\\(2.75)} &  \makecell{5.49\\(6.39)} & \makecell{3.67\\(3.47)} \\
             &       Test &  \makecell{85.0\\(21.18)} &   \makecell{62.13\\(32.4)} &  \makecell{89.37\\(8.82)} & \makecell{88.26\\(8.03)} & \makecell{89.24\\(8.46)} & \makecell{87.69\\(8.17)} &      \makecell{2.24\\(3.15)} &    \makecell{3.4\\(4.4)} & \makecell{2.42\\(2.74)} \\
             &        THT & \makecell{89.36\\(12.43)} &  \makecell{69.11\\(30.22)} & \makecell{90.04\\(10.02)} & \makecell{88.45\\(7.74)} &  \makecell{90.21\\(8.2)} & \makecell{87.44\\(8.42)} &      \makecell{3.59\\(3.66)} &    \makecell{3.9\\(4.9)} &  \makecell{3.18\\(2.9)} \\
\bottomrule
\end{tabular}
}
\caption{Results for High Generalization Performance policies.}
\label{tab:high_generalization_performance}
\end{table}

\begin{table}[htpb]
\centering
\resizebox{\columnwidth}{!}{
\begin{tabular}{cc|ccc|ccc|ccc|c}
\toprule
       &     & \multicolumn{3}{c|}{Individual} & \multicolumn{3}{c|}{Local} & \multicolumn{3}{c|}{Global} &  Summary \\
       &     &           Holdout &             Test &               THT &           Holdout &              Test &               THT &           Holdout &              Test &               THT & Test Accuracy $\uparrow$ \\
\midrule
\multirow{3}{*}{Individual} & Holdout &          $\equiv$ &  $\triangledown$ &   $\triangledown$ &          $\equiv$ &   $\triangledown$ &   $\triangledown$ &          $\equiv$ &   $\triangledown$ &   $\triangledown$ &       -6 \\
       & Test &  $\blacktriangle$ &         $\equiv$ &  $\blacktriangle$ &  $\blacktriangle$ &  $\blacktriangle$ &  $\blacktriangle$ &  $\blacktriangle$ &  $\blacktriangle$ &  $\blacktriangle$ &        8 \\
       & THT &  $\blacktriangle$ &  $\triangledown$ &          $\equiv$ &  $\blacktriangle$ &  $\blacktriangle$ &  $\blacktriangle$ &  $\blacktriangle$ &  $\blacktriangle$ &  $\blacktriangle$ &        6 \\
\midrule
\multirow{3}{*}{Local} & Holdout &          $\equiv$ &  $\triangledown$ &   $\triangledown$ &          $\equiv$ &   $\triangledown$ &   $\triangledown$ &          $\equiv$ &   $\triangledown$ &   $\triangledown$ &       -6 \\
       & Test &  $\blacktriangle$ &  $\triangledown$ &   $\triangledown$ &  $\blacktriangle$ &          $\equiv$ &  $\blacktriangle$ &  $\blacktriangle$ &  $\blacktriangle$ &  $\blacktriangle$ &        4 \\
       & THT &  $\blacktriangle$ &  $\triangledown$ &   $\triangledown$ &  $\blacktriangle$ &   $\triangledown$ &          $\equiv$ &  $\blacktriangle$ &  $\blacktriangle$ &  $\blacktriangle$ &        2 \\
\midrule
\multirow{3}{*}{Global} & Holdout &          $\equiv$ &  $\triangledown$ &   $\triangledown$ &          $\equiv$ &   $\triangledown$ &   $\triangledown$ &          $\equiv$ &   $\triangledown$ &   $\triangledown$ &       -6 \\
       & Test &  $\blacktriangle$ &  $\triangledown$ &   $\triangledown$ &  $\blacktriangle$ &   $\triangledown$ &   $\triangledown$ &  $\blacktriangle$ &          $\equiv$ &  $\blacktriangle$ &        0 \\
       & THT &  $\blacktriangle$ &  $\triangledown$ &   $\triangledown$ &  $\blacktriangle$ &   $\triangledown$ &   $\triangledown$ &  $\blacktriangle$ &   $\triangledown$ &          $\equiv$ &       -2 \\
\bottomrule

\end{tabular}
}
\caption{Wilcoxon Statistical Significance Test comparisons for High Generalization policies for different aggregations w.r.t. Test Accuracy.} 
\label{tab:statistical_high_generalization_performance}
\end{table}

We can analyze the architectural generalization by examining the three grouping strategies with the Test policy. The average number of neurons for Test policy with groupings Individual, Local and Global are $42.8, 68.5, 85$; while the accuracies are $90.61, 89.05, 87.69$, with Individual (win) -- Local, Local (win) -- Global. If we use Holdout instead of Test policy, we have the number of neurons $76.41, 86.41, 87.36$ and Test accuracy $85.54, 85.01, 85.48$. The Holdout Wilcoxon tests for the three groups are all equivalent. The Global grouping (for each dataset, we have chosen a single architecture) led to similar performances if we used the Local grouping (for each run, we choose the best model given the best-averaged architecture) or the Individual grouping (each run will have to treat each model individually, without any architecture grouping). However, Individual grouping could provide equivalent Test performances but with smaller networks.

It seems that analyzing the best architecture for a given data combination (the Individual grouping) is preferable. It contributes to our hypothesis that freezing the NN architecture is not the best approach. The models will experience different initializations and be exposed to slightly different training data. Therefore, they will experience different error landscapes, subjected to their learning dynamics.

We did not need to train several models to select good ones. The Individual--Holdout does not aggregate performances by the architecture, i.e., it does not use the concept of ``best architecture" for a given dataset, differently from the Local--Holdout and Global--Holdout. Still, the Individual--Holdout is performing equivalently to the Local--Holdout and Global--Holdout, according to Table~\ref{tab:statistical_high_generalization_performance}. We could select good models even when we have not used 10-fold cross-validation to aggregate architectures and decide the ``best architecture". Therefore, instead of repeating eight times the same architecture, we could always use different architectures using the same amount of memory and computational processing and increase the diversity of models by a factor of 7 (instead of $5,600/8=700$, we would have $5,600$ unique architectures). As we did not need to find the ``best architecture", we could use the time to run 10-fold cross-validation in other ways, such as varying non-architectural hyper-parameters, e.g., the learning rate or the maximum number of epochs.
Using different validation sets for the same architectures in a large pool of ANN might contribute to still finding good models. The training data change in the same run, and a specific combination of folds composing the training data might create better learning dynamics, producing exciting models.

When the Individual--Test policy is used, the Test accuracy is much higher than the Holdout accuracy. At the same time, the Train and Validation accuracies are minor if compared to the Test accuracy. Likewise, if the Individual--Holdout policy is used, the Test accuracy is smaller than the Holdout accuracy. Should we expect the model's performance in the real world to be closer to the Holdout or Test set? When we select the model that indiscriminately maximizes the Holdout or the Test metric (Holdout and Test policies), we are probably overestimating one metric and underestimating the other while potentially over-fitting the model. It can also be seen as a possible sign over-searching \citep{overfitting_overview_ying_2019}. Over-searching is a common problem when the hypothesis space of ML algorithms grows, increasing the over-fitting probability \citep{autoweka2013}. Over-searching is also a counterintuitive problem because we are trying to find the best hyper-parameters of the model, and the more points we know about the error landscape (more models trained and tested), the more we increase the chance of facing over-searching issues. In other words, the more information we have about our function to be optimized, the more we increase the chance of selecting a bad set of hyper-parameters. We argue that over-searching can be alleviated using a multi-criteria approach to select the models. 
We also advocate that the Disagreement between the Holdout-Test sets deserves attention when selecting or estimating the model's performance in the real world. 

From Table \ref{tab:statistical_high_generalization_performance}, we can realize that Individual--THT produces smaller NN and is better than both Local--THT and Global--THT. It also decreases the Disagreement between Holdout-Test and All if we compare them within the Individual group. It is worth mentioning that we have not tried to minimize the Disagreement measurements during the THT policy explicitly. However, this behavior had emerged while trying to maximize both Holdout and Test metrics.

Since the performance of the Individual--THT policy for Holdout and Test sets has the smallest Disagreement-Holdout-Test and also Disagreement--All, if compared to Individual--Holdout and Individual--Test policies, we could say that the real-world performance estimations would be more reliable than picking the model by individually maximizing the performance of each set using Individual--Holdout or Individual--Test (that is probably over-estimated in the optimized set and underestimated in the other).

It is also important to highlight that even though THT is deciding based on the Holdout and Test set metrics, the Individual--THT also increased both Train and Validation accuracy if compared against deciding only by maximizing the Test function. It contributes to our hypothesis that the models that have similar sets performance would be more robust, mentioned in section \ref{sec:optimality_condition_similar_sets_performance}.

Although seen as a bad practice due to the potential over-fitting, the Train policy is probably not catastrophic in our case due to the early stopping usage. Interestingly, the Train and Holdout approach contains very similar metrics. It might be because the Train set contains more data (70\%) than the Holdout set (10\%); therefore, it estimates better, and we are mitigating over-fitting by regularization and early stopping during training.

Since the results using Individual Aggregation are reasonable, we will not analyze the Local and Global aggregations in the subsequent results.

\subsection{High Robustness}
The results to understand the High Robustness Optimality Condition are presented in Table~\ref{tab:reduced_similar_sets_performance}. Also, the Wilcoxon Statistical Tests regarding the Test performance can be seen in Table~\ref{tab:reduced_statistical_similar_sets_performance}, and the Disagreement All is presented in Table~\ref{tab:reduced_statistical_similar_sets_performance_disagreement_all}. 

\begin{table}[htpb]
\centering
\resizebox{\columnwidth}{!}{
\begin{tabular}{c|c|c|cccc|ccc}
\toprule
 \multicolumn{1}{c|}{\multirow{2}{*}{Policy}}& \multirow{2}{*}{\# Neurons $\downarrow$}   & \multirow{2}{*}{\# Epochs $\uparrow$} &\multicolumn{4}{c|}{Accuracy $\uparrow$} & \multicolumn{3}{c}{Disagreement $\downarrow$} \\
                  &     &   &        Train &   Validation &      Holdout &         Test & Train-Validation & Holdout-Test &         All \\
\midrule
Holdout                      &   \makecell{76.41\\(27.5)} &  \makecell{84.79\\(25.13)} &  \makecell{92.08\\(7.92)} &  \makecell{89.94\\(7.79)} &  \makecell{91.76\\(7.66)} &  \makecell{85.54\\(9.29)} &      \makecell{2.77\\(2.66)} &  \makecell{6.37\\(6.53)} &  \makecell{3.94\\(3.51)} \\
TTVH                         &  \makecell{75.98\\(27.06)} &  \makecell{82.64\\(26.28)} &   \makecell{91.49\\(8.4)} &   \makecell{91.1\\(7.09)} &   \makecell{91.5\\(7.91)} &  \makecell{85.42\\(9.39)} &      \makecell{2.26\\(2.36)} &  \makecell{6.26\\(6.91)} &  \makecell{3.84\\(3.59)} \\
Test                         &   \makecell{42.8\\(33.91)} &  \makecell{63.72\\(34.72)} &   \makecell{87.42\\(8.8)} &  \makecell{87.07\\(7.64)} &  \makecell{87.68\\(7.98)} &  \makecell{90.61\\(7.05)} &      \makecell{2.67\\(3.25)} &  \makecell{4.29\\(3.51)} &  \makecell{3.13\\(2.41)} \\
THT                          &   \makecell{75.5\\(24.61)} &  \makecell{78.21\\(27.19)} &  \makecell{91.22\\(8.36)} &  \makecell{89.05\\(7.89)} &   \makecell{90.9\\(8.07)} &  \makecell{89.65\\(7.34)} &      \makecell{2.83\\(2.67)} &  \makecell{3.52\\(3.26)} &   \makecell{2.67\\(2.0)} \\
TTVHT                        &   \makecell{76.5\\(24.32)} &  \makecell{77.75\\(27.45)} &   \makecell{91.05\\(8.5)} &  \makecell{90.47\\(7.21)} &   \makecell{91.05\\(8.0)} &  \makecell{88.88\\(7.78)} &      \makecell{2.52\\(2.45)} &  \makecell{3.42\\(3.53)} &  \makecell{2.48\\(2.11)} \\
\bottomrule
\end{tabular}
}
\caption{Results for High Robustness policies.}
\label{tab:reduced_similar_sets_performance}
\end{table}

\begin{table}[htpb]
\centering
\resizebox{\columnwidth}{!}{

\begin{tabular}{c|ccccc|c}
\toprule
Policy &           Holdout &              TTVH &             Test &               THT &             TTVHT &  Summary Test Accuracy $\uparrow$ \\
\midrule
Holdout &          $\equiv$ &          $\equiv$ &  $\triangledown$ &   $\triangledown$ &   $\triangledown$ &                -3 \\
TTVH    &          $\equiv$ &          $\equiv$ &  $\triangledown$ &   $\triangledown$ &   $\triangledown$ &                -3 \\
Test    &  $\blacktriangle$ &  $\blacktriangle$ &         $\equiv$ &  $\blacktriangle$ &  $\blacktriangle$ &                 4 \\
THT     &  $\blacktriangle$ &  $\blacktriangle$ &  $\triangledown$ &          $\equiv$ &  $\blacktriangle$ &                 2 \\
TTVHT   &  $\blacktriangle$ &  $\blacktriangle$ &  $\triangledown$ &   $\triangledown$ &          $\equiv$ &                 0 \\
\bottomrule
\end{tabular}
}
\caption{Wilcoxon Statistical Significance Test comparisons for High Robustness policies w.r.t. Test Accuracy.}
\label{tab:reduced_statistical_similar_sets_performance}
\end{table}

\begin{table}[htpb]
\centering
\resizebox{\columnwidth}{!}{
\begin{tabular}{c|ccccc|c}
\toprule
Policy &          Holdout &              TTVH &              Test &               THT &             TTVHT &  Summary All Disagreement $\downarrow$ \\
\midrule
Holdout &         $\equiv$ &  $\blacktriangle$ &  $\blacktriangle$ &  $\blacktriangle$ &  $\blacktriangle$ &                         4 \\
TTVH    &  $\triangledown$ &          $\equiv$ &          $\equiv$ &  $\blacktriangle$ &  $\blacktriangle$ &                         1 \\
Test    &  $\triangledown$ &          $\equiv$ &          $\equiv$ &  $\blacktriangle$ &  $\blacktriangle$ &                         1 \\
THT     &  $\triangledown$ &   $\triangledown$ &   $\triangledown$ &          $\equiv$ &  $\blacktriangle$ &                        -2 \\
TTVHT   &  $\triangledown$ &   $\triangledown$ &   $\triangledown$ &   $\triangledown$ &          $\equiv$ &                        -4 \\
\bottomrule
\end{tabular}
}
\caption{Wilcoxon Statistical Significance Test comparisons for High Robustness policies w.r.t. All Disagreement.}
\label{tab:reduced_statistical_similar_sets_performance_disagreement_all}
\end{table}

We argue that the difference of the performance metrics for each set is inversely proportional to its robustness. Let us compare the policies that do not look into the Test set (TTVH and Holdout). We can see that TTVH could select models with statistically equivalent results regarding the Test performance, as depicted in Table~\ref{tab:reduced_statistical_similar_sets_performance}. At the same time, from Table~\ref{tab:reduced_statistical_similar_sets_performance_disagreement_all} it was able to significantly decrease the Disagreement if compared to the Holdout policy. Therefore, instead of looking only into the Holdout performance metric, we should use the TTVH approach since it produces statistically equivalent performance metrics and still decreases the Disagreement between the sets' performances. This probably indicates a more robust selection.

We have generally been taught that the best model is the one that maximizes the Test performance metric. This is evident if we consider how state-of-the-art methods are traditionally benchmarked: very often only analyzing if the Test performance metric is better than previous methods applied to the same dataset. We argue that the performance metrics of the other sets must also be involved for a complete evaluation. If we recall the Equation~\ref{eq:data_explanation_decomposition_test}, $Y_{test}=M(X_{test})+\epsilon_{test}$, it might be the case that we select the model that maximizes $M(X_{test}) + \epsilon_{test}$ on the Test data, providing a probably over-estimation for the model performance during production, instead of what we really want: the model that maximizes only the $M(X_{test})$ term, since when using the model in production, we will probably have a $\epsilon_{production}$ different from our $\epsilon_{test}$.

Comparing the policies that include the Test metric in the decision, the Test policy should be the upper bound policy regarding the Test metric if we use the . However, as we mentioned, it is probably over-estimated due to inherent noise in the Test set. Therefore we need to be cautious about taking this as the desired model. To corroborate this idea, we can see from the THT policy that the Train and Validation metrics improved compared to the Test policy. However, we have not explicitly optimized for that. On the other hand, the TTVHT model (which also optimizes the Train and Validation metrics) would probably be our target model since it tries to maximize the performance over the Train, Validation, Holdout, and Test sets. As a consequence of the better performance equilibrium in the TTVHT and THT compared to the Test policy, we can also see that the Disagreement was significantly decreased in Table~\ref{tab:reduced_statistical_similar_sets_performance_disagreement_all}.

If we compare the TTVHT to the TTVH approach, we can see that the Number of Neurons, Train, Validation, and Holdout set performances are similar. Let us consider only the subset of models which deliver similar performance to the TTVH values (similar to freezing the previously mentioned metrics). We have a group of models that varies the Test metric. The TTVHT would be an approximation to this group's best model, but TTVH could not select it. It probably means that we can include other attributes that describe the learning dynamics of the model in the TOPSIS decision process, such as the difference between the initial and final performances and weight regularization values, to create better model selectors. It opens many possibilities to improve the multi-criteria model selection procedure.

\subsection{Low Complexity}
To analyze the Low Complexity Optimality Condition, we are trying to minimize the number of neurons in a model. The results for each policy containing and without the Number of Neurons as decision criteria are presented in Table~\ref{tab:reduced_low_complexity}. The Wilcoxon statistical test results of the same set of policies regarding the Test and Disagreement are respectively presented in Table~\ref{tab:reduced_statistical_low_complexity} and Table~\ref{tab:reduced_statistical_low_complexity_disagreement_all}
\begin{table}[htpb]
\centering
\resizebox{\columnwidth}{!}{
\begin{tabular}{c|c|c|cccc|ccc}
\toprule
 \multicolumn{1}{c|}{\multirow{2}{*}{Policy}}& \multirow{2}{*}{\# Neurons $\downarrow$}   & \multirow{2}{*}{\# Epochs $\uparrow$} &\multicolumn{4}{c|}{Accuracy $\uparrow$} & \multicolumn{3}{c}{Disagreement $\downarrow$} \\
                  &     &   &        Train &   Validation &      Holdout &         Test & Train-Validation & Holdout-Test &         All \\
\midrule
TTVH                         &  \makecell{75.98\\(27.06)} &  \makecell{82.64\\(26.28)} &   \makecell{91.49\\(8.4)} &   \makecell{91.1\\(7.09)} &   \makecell{91.5\\(7.91)} &  \makecell{85.42\\(9.39)} &      \makecell{2.26\\(2.36)} &  \makecell{6.26\\(6.91)} &  \makecell{3.84\\(3.59)} \\
TTVHN                        &   \makecell{10.27\\(9.27)} &   \makecell{78.08\\(28.3)} &  \makecell{87.59\\(7.94)} &  \makecell{88.36\\(7.13)} &  \makecell{87.83\\(7.66)} &   \makecell{82.9\\(9.96)} &      \makecell{2.14\\(1.93)} &  \makecell{5.54\\(7.19)} &  \makecell{3.54\\(3.74)} \\
THT                          &   \makecell{75.5\\(24.61)} &  \makecell{78.21\\(27.19)} &  \makecell{91.22\\(8.36)} &  \makecell{89.05\\(7.89)} &   \makecell{90.9\\(8.07)} &  \makecell{89.65\\(7.34)} &      \makecell{2.83\\(2.67)} &  \makecell{3.52\\(3.26)} &   \makecell{2.67\\(2.0)} \\
THTN                         &    \makecell{8.75\\(7.24)} &   \makecell{78.35\\(27.0)} &  \makecell{86.78\\(7.93)} &  \makecell{85.97\\(8.22)} &    \makecell{86.7\\(7.8)} &   \makecell{86.9\\(7.94)} &      \makecell{2.37\\(2.65)} &  \makecell{2.75\\(2.11)} &    \makecell{2.2\\(1.6)} \\
TTVHT                        &   \makecell{76.5\\(24.32)} &  \makecell{77.75\\(27.45)} &   \makecell{91.05\\(8.5)} &  \makecell{90.47\\(7.21)} &   \makecell{91.05\\(8.0)} &  \makecell{88.88\\(7.78)} &      \makecell{2.52\\(2.45)} &  \makecell{3.42\\(3.53)} &  \makecell{2.48\\(2.11)} \\
TTVHTN                       &   \makecell{11.37\\(9.44)} &  \makecell{77.48\\(27.79)} &  \makecell{87.54\\(7.92)} &  \makecell{88.14\\(7.13)} &  \makecell{87.76\\(7.61)} &  \makecell{86.44\\(8.35)} &      \makecell{2.09\\(1.93)} &  \makecell{3.15\\(3.58)} &   \makecell{2.3\\(1.93)} \\
\bottomrule
\end{tabular}
}
\caption{Results for Low Complexity policies.}
\label{tab:reduced_low_complexity}
\end{table}

\begin{table}[htpb]
\centering
\resizebox{\columnwidth}{!}{
\begin{tabular}{c|cccccc|c}
\toprule
Policy &              TTVH &             TTVHN &              THT &              THTN &             TTVHT &            TTVHTN &  Summary Test Accuracy $\uparrow$ \\
\midrule
TTVH   &          $\equiv$ &  $\blacktriangle$ &  $\triangledown$ &   $\triangledown$ &   $\triangledown$ &   $\triangledown$ &                -3 \\
TTVHN  &   $\triangledown$ &          $\equiv$ &  $\triangledown$ &   $\triangledown$ &   $\triangledown$ &   $\triangledown$ &                -5 \\
THT    &  $\blacktriangle$ &  $\blacktriangle$ &         $\equiv$ &  $\blacktriangle$ &  $\blacktriangle$ &  $\blacktriangle$ &                 5 \\
THTN   &  $\blacktriangle$ &  $\blacktriangle$ &  $\triangledown$ &          $\equiv$ &   $\triangledown$ &  $\blacktriangle$ &                 1 \\
TTVHT  &  $\blacktriangle$ &  $\blacktriangle$ &  $\triangledown$ &  $\blacktriangle$ &          $\equiv$ &  $\blacktriangle$ &                 3 \\
TTVHTN &  $\blacktriangle$ &  $\blacktriangle$ &  $\triangledown$ &   $\triangledown$ &   $\triangledown$ &          $\equiv$ &                -1 \\
\bottomrule
\end{tabular}
}
\caption{Wilcoxon Statistical Significance Test comparisons for Low Complexity policies w.r.t. Test Accuracy.}
\label{tab:reduced_statistical_low_complexity}

\end{table}

\begin{table}[htpb]
\centering

\resizebox{\columnwidth}{!}{
\begin{tabular}{c|cccccc|c}
\toprule
Policy &              TTVH &             TTVHN &              THT &              THTN &             TTVHT &            TTVHTN &  Summary All Disagreement $\downarrow$ \\
\midrule
TTVH   &         $\equiv$ &  $\blacktriangle$ &  $\blacktriangle$ &  $\blacktriangle$ &  $\blacktriangle$ &  $\blacktriangle$ &                         5 \\
TTVHN  &  $\triangledown$ &          $\equiv$ &  $\blacktriangle$ &  $\blacktriangle$ &  $\blacktriangle$ &  $\blacktriangle$ &                         3 \\
THT    &  $\triangledown$ &   $\triangledown$ &          $\equiv$ &  $\blacktriangle$ &  $\blacktriangle$ &  $\blacktriangle$ &                         1 \\
THTN   &  $\triangledown$ &   $\triangledown$ &   $\triangledown$ &          $\equiv$ &   $\triangledown$ &          $\equiv$ &                        -4 \\
TTVHT  &  $\triangledown$ &   $\triangledown$ &   $\triangledown$ &  $\blacktriangle$ &          $\equiv$ &  $\blacktriangle$ &                        -1 \\
TTVHTN &  $\triangledown$ &   $\triangledown$ &   $\triangledown$ &          $\equiv$ &   $\triangledown$ &          $\equiv$ &                        -4 \\
\bottomrule
\end{tabular}
}
\caption{Wilcoxon Statistical Significance Test comparisons for Low Complexity policies w.r.t. All Disagreement.}
\label{tab:reduced_statistical_low_complexity_disagreement_all}
\end{table}

The models decreased the average accuracy when adding the number of neurons in the multi-criteria equation. However, it largely dropped the number of neurons necessary to encode the knowledge of each dataset. It is essential to create simpler models since they tend to generalize better because they make fewer assumptions about the data it is trying to learn.

It is worth mentioning that TOPSIS allows us to set different weights for each objective, even though we have used the same weight for all objectives. Tuning this specific TOPSIS parameter will lead to different results based on the importance we are willing to give for each objective.


\subsection{No Premature Early Stopping}
We have included the number of epochs in the TOPSIS decision-making process to analyze the importance of the Early Stopping epoch. The ranking policies end with E (stopping close to the End of training) or B (stopping close to the Begin of the training). 

As we can see from Table~\ref{tab:reduced_no_premature_early_stopping}, the ranking policies that maximize the number of trained epochs (ends with E) consistently outperform their counterparts that minimize the number of trained epochs (ends with B) for all the Aggregation and Ranking policies. It shows that how long the model was trained is also important to consider during the model selection.
Even though when directly comparing TTVHN vs. TTVHNE (that includes the maximization of trained epochs), we can see from Table~\ref{tab:reduced_statistical_no_premature_early_stopping} that the performance on the Test set is equivalent as well as the Disagreement is not significantly different. 
Since no statistically better results were found by maximizing the Number of trained Epochs, but it still affects the model selection when comparing B vs. E policies, it might be the case that we need to tweak better the weight for this specific criteria to mostly avoid premature early stopping (B) instead of looking for late or no early stopping at all (E), which can also contain models that have not converged yet at the final epochs. This probably would be better used as a filter to avoid premature early stopping (B), but not be simultaneously maximized.


\begin{table}[htpb]
\centering
\resizebox{\columnwidth}{!}{
\begin{tabular}{c|c|c|cccc|ccc}
\toprule
 \multicolumn{1}{c|}{\multirow{2}{*}{Policy}}& \multirow{2}{*}{\# Neurons $\downarrow$}   & \multirow{2}{*}{\# Epochs $\uparrow$} &\multicolumn{4}{c|}{Accuracy $\uparrow$} & \multicolumn{3}{c}{Disagreement $\downarrow$} \\
                  &     &   &        Train &   Validation &      Holdout &         Test & Train-Validation & Holdout-Test &         All \\
\midrule
TTVHN                        &   \makecell{10.27\\(9.27)} &   \makecell{78.08\\(28.3)} &  \makecell{87.59\\(7.94)} &  \makecell{88.36\\(7.13)} &  \makecell{87.83\\(7.66)} &    \makecell{82.9\\(9.96)} &      \makecell{2.14\\(1.93)} &  \makecell{5.54\\(7.19)} &  \makecell{3.54\\(3.74)} \\
TTVHNB                       &   \makecell{13.8\\(12.23)} &  \makecell{17.94\\(20.68)} &  \makecell{83.41\\(7.61)} &  \makecell{85.37\\(7.33)} &  \makecell{84.01\\(7.38)} &  \makecell{80.48\\(10.26)} &      \makecell{2.68\\(2.17)} &  \makecell{4.63\\(5.87)} &  \makecell{3.34\\(3.24)} \\
TTVHNE                       &   \makecell{10.77\\(9.38)} &  \makecell{89.83\\(24.46)} &  \makecell{87.53\\(7.96)} &  \makecell{87.79\\(7.08)} &  \makecell{87.66\\(7.66)} &   \makecell{82.96\\(9.91)} &       \makecell{2.2\\(2.06)} &  \makecell{5.33\\(6.87)} &  \makecell{3.41\\(3.64)} \\
THTN                         &    \makecell{8.75\\(7.24)} &   \makecell{78.35\\(27.0)} &  \makecell{86.78\\(7.93)} &  \makecell{85.97\\(8.22)} &    \makecell{86.7\\(7.8)} &    \makecell{86.9\\(7.94)} &      \makecell{2.37\\(2.65)} &  \makecell{2.75\\(2.11)} &    \makecell{2.2\\(1.6)} \\
THTNB                        &   \makecell{11.17\\(8.83)} &   \makecell{16.53\\(20.6)} &  \makecell{81.83\\(8.28)} &  \makecell{83.49\\(8.61)} &  \makecell{82.49\\(8.14)} &   \makecell{82.63\\(9.74)} &       \makecell{2.39\\(2.7)} &  \makecell{3.08\\(3.03)} &  \makecell{2.44\\(2.04)} \\
THTNE                        &     \makecell{8.49\\(6.9)} &  \makecell{89.79\\(24.18)} &  \makecell{86.69\\(8.26)} &  \makecell{85.53\\(8.19)} &  \makecell{86.61\\(7.89)} &   \makecell{86.14\\(8.12)} &      \makecell{2.53\\(2.84)} &  \makecell{2.62\\(2.69)} &  \makecell{2.19\\(1.88)} \\
TTVHTN                       &   \makecell{11.37\\(9.44)} &  \makecell{77.48\\(27.79)} &  \makecell{87.54\\(7.92)} &  \makecell{88.14\\(7.13)} &  \makecell{87.76\\(7.61)} &   \makecell{86.44\\(8.35)} &      \makecell{2.09\\(1.93)} &  \makecell{3.15\\(3.58)} &   \makecell{2.3\\(1.93)} \\
TTVHTNE                      &  \makecell{11.89\\(10.47)} &  \makecell{89.55\\(24.41)} &  \makecell{87.53\\(7.93)} &  \makecell{87.85\\(7.19)} &   \makecell{87.7\\(7.66)} &     \makecell{85.9\\(8.3)} &      \makecell{2.04\\(2.11)} &  \makecell{3.39\\(4.05)} &  \makecell{2.39\\(2.23)} \\
TTVHTNB                      &  \makecell{14.91\\(12.97)} &  \makecell{19.09\\(21.61)} &  \makecell{83.78\\(7.08)} &   \makecell{85.7\\(6.92)} &  \makecell{84.38\\(6.89)} &   \makecell{83.37\\(8.09)} &       \makecell{2.59\\(2.4)} &   \makecell{3.3\\(3.41)} &  \makecell{2.62\\(2.04)} \\
\bottomrule
\end{tabular}
}
\caption{Results for No Premature Early Stopping policies.}
\label{tab:reduced_no_premature_early_stopping}

\end{table}

\begin{table}[htpb]
\centering
\resizebox{\columnwidth}{!}{
\begin{tabular}{c|ccccccccc|c}
\toprule
Policy &             TTVHN &            TTVHNB &            TTVHNE &             THTN &             THTNB &             THTNE &            TTVHTN &           TTVHTNE &           TTVHTNB &  Summary Test Accuracy $\uparrow$ \\
\midrule
TTVHN   &          $\equiv$ &  $\blacktriangle$ &          $\equiv$ &  $\triangledown$ &          $\equiv$ &   $\triangledown$ &   $\triangledown$ &   $\triangledown$ &          $\equiv$ &                -3 \\
TTVHNB  &   $\triangledown$ &          $\equiv$ &   $\triangledown$ &  $\triangledown$ &   $\triangledown$ &   $\triangledown$ &   $\triangledown$ &   $\triangledown$ &   $\triangledown$ &                -8 \\
TTVHNE  &          $\equiv$ &  $\blacktriangle$ &          $\equiv$ &  $\triangledown$ &          $\equiv$ &   $\triangledown$ &   $\triangledown$ &   $\triangledown$ &          $\equiv$ &                -3 \\
THTN    &  $\blacktriangle$ &  $\blacktriangle$ &  $\blacktriangle$ &         $\equiv$ &  $\blacktriangle$ &  $\blacktriangle$ &  $\blacktriangle$ &  $\blacktriangle$ &  $\blacktriangle$ &                 8 \\
THTNB   &          $\equiv$ &  $\blacktriangle$ &          $\equiv$ &  $\triangledown$ &          $\equiv$ &   $\triangledown$ &   $\triangledown$ &   $\triangledown$ &   $\triangledown$ &                -4 \\
THTNE   &  $\blacktriangle$ &  $\blacktriangle$ &  $\blacktriangle$ &  $\triangledown$ &  $\blacktriangle$ &          $\equiv$ &   $\triangledown$ &          $\equiv$ &  $\blacktriangle$ &                 3 \\
TTVHTN  &  $\blacktriangle$ &  $\blacktriangle$ &  $\blacktriangle$ &  $\triangledown$ &  $\blacktriangle$ &  $\blacktriangle$ &          $\equiv$ &  $\blacktriangle$ &  $\blacktriangle$ &                 6 \\
TTVHTNE &  $\blacktriangle$ &  $\blacktriangle$ &  $\blacktriangle$ &  $\triangledown$ &  $\blacktriangle$ &          $\equiv$ &   $\triangledown$ &          $\equiv$ &  $\blacktriangle$ &                 3 \\
TTVHTNB &          $\equiv$ &  $\blacktriangle$ &          $\equiv$ &  $\triangledown$ &  $\blacktriangle$ &   $\triangledown$ &   $\triangledown$ &   $\triangledown$ &          $\equiv$ &                -2 \\
\bottomrule
\end{tabular}
}
\caption{Wilcoxon Statistical Significance Test comparisons for No Premature Early Stopping policies w.r.t. Test Accuracy.}
\label{tab:reduced_statistical_no_premature_early_stopping}

\end{table}

\begin{table}[htpb]
\centering
\resizebox{\columnwidth}{!}{
\begin{tabular}{c|ccccccccc|c}
\toprule
Policy &             TTVHN &            TTVHNB &            TTVHNE &             THTN &             THTNB &             THTNE &            TTVHTN &           TTVHTNE &           TTVHTNB &  Summary All Disagreement $\downarrow$ \\
\midrule
TTVHN   &         $\equiv$ &         $\equiv$ &         $\equiv$ &  $\blacktriangle$ &  $\blacktriangle$ &  $\blacktriangle$ &  $\blacktriangle$ &  $\blacktriangle$ &  $\blacktriangle$ &                         6 \\
TTVHNB  &         $\equiv$ &         $\equiv$ &         $\equiv$ &  $\blacktriangle$ &  $\blacktriangle$ &  $\blacktriangle$ &  $\blacktriangle$ &  $\blacktriangle$ &  $\blacktriangle$ &                         6 \\
TTVHNE  &         $\equiv$ &         $\equiv$ &         $\equiv$ &  $\blacktriangle$ &  $\blacktriangle$ &  $\blacktriangle$ &  $\blacktriangle$ &  $\blacktriangle$ &  $\blacktriangle$ &                         6 \\
THTN    &  $\triangledown$ &  $\triangledown$ &  $\triangledown$ &          $\equiv$ &          $\equiv$ &          $\equiv$ &          $\equiv$ &          $\equiv$ &   $\triangledown$ &                        -4 \\
THTNB   &  $\triangledown$ &  $\triangledown$ &  $\triangledown$ &          $\equiv$ &          $\equiv$ &  $\blacktriangle$ &          $\equiv$ &          $\equiv$ &   $\triangledown$ &                        -3 \\
THTNE   &  $\triangledown$ &  $\triangledown$ &  $\triangledown$ &          $\equiv$ &   $\triangledown$ &          $\equiv$ &          $\equiv$ &          $\equiv$ &   $\triangledown$ &                        -5 \\
TTVHTN  &  $\triangledown$ &  $\triangledown$ &  $\triangledown$ &          $\equiv$ &          $\equiv$ &          $\equiv$ &          $\equiv$ &          $\equiv$ &   $\triangledown$ &                        -4 \\
TTVHTNE &  $\triangledown$ &  $\triangledown$ &  $\triangledown$ &          $\equiv$ &          $\equiv$ &          $\equiv$ &          $\equiv$ &          $\equiv$ &   $\triangledown$ &                        -4 \\
TTVHTNB &  $\triangledown$ &  $\triangledown$ &  $\triangledown$ &  $\blacktriangle$ &  $\blacktriangle$ &  $\blacktriangle$ &  $\blacktriangle$ &  $\blacktriangle$ &          $\equiv$ &                         2 \\
\bottomrule
\end{tabular}
}
\caption{Wilcoxon Statistical Significance Test comparisons for No Premature Early Stopping policies w.r.t. All Disagreement.}
\label{tab:reduced_statistical_no_premature_early_stopping_disagreement_all}

\end{table}

\section{Conclusions}
\label{sec:conclusions}
In this work, we propose to use several criteria of a machine learning model, in this specific case, Neural Networks, to perform model selection (multi-criteria model selection) instead of using the most common approach of a single criterion. 
We have shown that fixing the number of Neurons of an ANN, primarily done in ANN training methodology, for a specific dataset does not produce the best possible models. Using a flexible architecture usually leads to better results. Also, we empirically demonstrated that over-fitting and over-searching could be mitigated by performing a multi-criteria model selection procedure instead of a single criterion when deciding from a very large pool of candidates because it alleviates the metrics maximization on a specific set that the model usually is also explaining the noise. When a multi-criteria approach is used, we are probably diluting the risk of noise fitting through all the sets since we expect the noise should not be modeled, and using more data from different sets that agree with each other during the model selection, we increase our confidence on the model assessment or performance estimation.

We have also defined optimality conditions that we desire to have in theoretical models and empirically demonstrated that those conditions seem important during model selection.
The oracle or any model selected solely based on a single performance for a set is usually over-optimistic and probably over-fitted on that specific set due to the over-searching issue and deserves attention if one would consider them as the best targets during model selection.
We believe this work might lead to different research directions since it gives a different perspective and justifies why model selection that only maximizes specific performance metrics tends not to be the best approach.

\section{Future Works}
\label{sec:future_works}
We would like to evaluate our policies within AutoML algorithms, which usually only perform model selection based on a single objective. In this process, it might also be interesting to build deep neural networks layer-by-layer using ParallelMLPs to train several candidates and use TOPSIS to rank and choose the current layer, appending layer by layer. Using other learning dynamic attributes and tweaking the weights for each criterion might produce better model selections.
We intend to try different MCDM methods with different variables and weights and vector normalization schemes to investigate if we can select better models.
In order to collect more evidence on our optimality conditions, we would like to analyze adversarial attack influence for models using single-criterion and multi-criteria model selections.
With respect to the No Premature Early Stopping condition, a further investigation is needed since it has influence on the decision process but apparently we were not able to fully capture it. This probably can be converted into a filter to avoid premature early stopped models, but without emphasizing longer training, since good models can still have good performance but converging faster than other late converged models.


\acks{We would like to acknowledge the Centro de Tecnologias Estratégicas do Nordeste (CETENE) for providing computational resources (FACEPE APQ-1864-1.06/12) and to thank FACEPE, CNPq and CAPES (Brazilian Research Agencies) for their financial support.
}










\vskip 0.2in
\bibliography{main}

\end{document}